\documentclass[conference,a4paper]{IEEEtran}
\IEEEoverridecommandlockouts

\usepackage{cite}
\usepackage{amsmath,amssymb,amsfonts}
\usepackage{graphicx}
\usepackage[caption=false,font=footnotesize]{subfig}
\usepackage{textcomp}
\usepackage{xcolor}
\usepackage{array}
\usepackage{pifont}
\newcommand{\cmark}{\ding{51}}
\newcommand{\xmark}{\ding{55}}
\usepackage{url}
\usepackage{eso-pic}
\def\BibTeX{{\rm B\kern-.05em{\sc i\kern-.025em b}\kern-.08em
    T\kern-.1667em\lower.7ex\hbox{E}\kern-.125emX}}
\begin{document}
\AddToShipoutPictureFG*{\AtPageLowerLeft{%
  \put(\LenToUnit{\dimexpr(\paperwidth-\textwidth)/2\relax},\LenToUnit{48pt}){%
    \parbox[b]{\textwidth}{\centering\scriptsize
      Accepted for publication at the International Conference on Digital Image
      Computing: Techniques and Applications (DICTA), 2026.\\
      \copyright~2026 IEEE. Personal use of this material is permitted. Permission
      from IEEE must be obtained for all other uses, in any current or future media,
      including reprinting/republishing this material for advertising or promotional
      purposes, creating new collective works, for resale or redistribution to servers
      or lists, or reuse of any copyrighted component of this work in other works.}}}}

\title{Squeeze-Aware-YOLO: Early Channel Recalibration and Multi-Scale Fusion for Real-Time Traffic Sign Detection}

\author{\IEEEauthorblockN{Mizbaul Haque Maruf, Syed Samin Sadaf, Syed Tamzid Bakht,
Md. Hasanul Kabir, Sabbir Ahmed}
\IEEEauthorblockA{\textit{Dept. of Computer Science and Engineering, Islamic University of Technology}\\
Boardbazar, Gazipur, Bangladesh \\
\{mizbaulhaque, saminsadaf, syedtamzid, hasanul, sabbirahmed\}@iut-dhaka.edu}
}

\maketitle

\begin{abstract}
Traffic sign detection plays a vital role in intelligent transportation systems, supporting the safe operation of advanced driver assistance systems and autonomous vehicles.
Despite significant progress in real-time object detection, accurately identifying traffic signs under challenging conditions remains difficult. Variations in lighting, adverse weather, motion blur, and partial occlusion often reduce detection performance, especially for small or visually degraded signs where important spatial information can be lost.
To address these challenges, we propose Squeeze-Aware-YOLO, a lightweight traffic sign detection framework that introduces a multi-branch bottleneck module combining convolutional kernels of different sizes with a Squeeze-and-Excitation mechanism to improve feature extraction and adaptive channel weighting.
In addition, we integrate the Multi-Branch Auxiliary Feature Pyramid Network and Re-parameterized Heterogeneous Multi-Scale modules to strengthen multi-scale feature representation and feature fusion while preserving real-time inference efficiency for practical deployment.
The proposed approach is evaluated on the Bangladeshi Traffic Sign Detection Dataset, the CSUST Chinese Traffic Sign Detection Benchmark 2021, and the German Traffic Sign Detection Benchmark, achieving mAP@0.5 scores of 74.7\%, 79.4\%, and 84.0\%, respectively, while consistently outperforming recent state-of-the-art YOLO-based traffic sign detectors.
These findings demonstrate that combining multi-scale receptive fields, channel attention, and enhanced feature fusion enables more accurate and robust traffic sign detection while maintaining the computational efficiency required for real-world deployment.
\end{abstract}


\section{Introduction}
Object detection is a fundamental task in computer vision that simultaneously identifies object categories and accurately localizes them within an image using bounding boxes.
Since the introduction of deep learning, detection and recognition systems have achieved significant success across a wide range of applications~\cite{sultana2020review}.
Among these, traffic sign detection is a critical component of autonomous driving systems and advanced driver-assistance systems.
Accurate traffic sign detection directly enhances road safety by enabling vehicles to reliably perceive and interpret visual cues in real time, even under dynamic conditions~\cite{zhu2016traffic,wali2019vision}.

Although traffic sign detection has evolved from hand-crafted features to transformer architectures, many approaches remain dataset-dependent and lose accuracy under illumination variation, occlusion, complex backgrounds, and changes in sign appearance~\cite{wali2019vision,choudhary2026faierdet,gray2023glare}, while transformer-based methods incur computational costs that limit real-time deployment~\cite{yu2026blocked}.
These limitations are more pronounced in countries such as Bangladesh, where non-standard signage, dense traffic, faded or damaged signs, poor illumination, and adverse weather pose additional challenges~\cite{ashik2022recognizing}.

Established benchmarks such as GTSDB~\cite{houben2013detection}, Tsinghua-Tencent 100K~\cite{zhu2016traffic}, and LISA~\cite{mogelmose2012vision} have advanced the field, but they are relatively controlled, and traffic signs typically occupy only about 0.2\% of a road scene, leaving little visual information to exploit~\cite{zhu2016traffic}.
Compared with computationally intensive transformer-based models~\cite{manzari2022pyramid}, YOLO-based detectors jointly perform localization and classification in a lightweight single-stage architecture, balancing accuracy and inference speed for embedded intelligent transportation systems~\cite{redmon2016you}.
Nevertheless, they still struggle to reliably detect small, degraded, and densely occluded signs in complex real-world scenes~\cite{chu2023trd,du2024tsd}.

To address these challenges, we propose Squeeze-Aware-YOLO, a robust traffic sign detection framework that improves feature representation for small and visually degraded objects while preserving the real-time inference capability required for intelligent transportation systems. The main contributions of this work are as follows:
\begin{itemize}
    \item A novel squeeze-and-excitation (SE)~\cite{hu2018squeeze} enhanced triple-bottleneck feature extraction module that seamlessly integrates adaptive channel recalibration with progressive multi-scale representation learning.
    \item A unified architectural enhancement that synergistically combines re-parameterized hierarchical feature extraction with efficient multi-scale feature fusion, facilitating richer information exchange across semantic levels.
    \item Evaluations on the Bangladeshi Traffic Sign Dataset (BTSD), the CSUST Chinese Traffic Sign Detection Benchmark 2021 (CCTSDB2021), and the German Traffic Sign Detection Benchmark (GTSDB) demonstrate consistent effectiveness across three independently trained benchmarks spanning distinct geographic regions, sign taxonomies, and capture conditions, together with a per-class failure analysis that characterizes where the gains and regressions occur.
\end{itemize}

\section{Related Works}\label{sec:literature}
Since its introduction, YOLO (You Only Look Once) has evolved through continuous architectural refinement.
The original YOLO~\cite{redmon2016you} established the foundation for single-stage detection but struggled with small and densely distributed objects.
Later versions introduced deeper backbones, multi-scale prediction, feature aggregation, and improved training strategies~\cite{redmon2017yolo9000,redmon2018yolov3,bochkovskiy2020yolov4}, while YOLOv8--YOLOv11 further balanced accuracy and inference speed through anchor-free detection, structural re-parameterization, and attention mechanisms~\cite{jocher2023yolov8,wang2024yolov10,jocher2024yolo11,wang2024yolov1,terven2023comprehensive}.
Reliable detection of small, densely distributed, and partially occluded objects under real-time constraints nonetheless remains a significant challenge~\cite{flores2024traffic}.

Progress in traffic sign detection has been closely tied to increasingly diverse benchmarks. Early datasets such as the Belgium~\cite{timofte2014multi} and Swedish~\cite{larsson2011using} traffic sign datasets and GTSDB~\cite{houben2013detection} standardized evaluation but offered limited environmental diversity, whereas LISA~\cite{mogelmose2012vision}, TT100K~\cite{zhu2016traffic}, Mapillary~\cite{ertler2020mapillary}, and region-specific sets from Malaysia~\cite{madani2016malaysian}, India~\cite{megalingam2023indian}, and Bangladesh (BTSD)~\cite{ashik2022recognizing} added variation in weather, illumination, traffic density, and sign appearance. Methodologically, early detectors relied on color segmentation, shape analysis, HOG, and Haar-like descriptors~\cite{berkaya2016circular,salti2015traffic} and were highly sensitive to illumination changes, occlusion, and clutter, whereas deep detectors such as Faster R-CNN~\cite{ren2017faster}, SSD~\cite{liu2016ssd}, and the YOLO family~\cite{redmon2016you} learn hierarchical representations directly from data.

Recent studies have focused on multi-scale feature extraction and contextual representation. Chu \textit{et al.}~\cite{chu2023trd} introduced self-attention to capture global context, and Liu \textit{et al.}~\cite{liu2023etsr} paired Coordinate Attention with ConvNeXt blocks to strengthen feature extraction. Du \textit{et al.}~\cite{du2024tsd} used Space-to-Depth transformation and Selective Kernel attention, at the expense of increased architectural complexity. In challenging road environments such as Bangladesh, Ashik \textit{et al.}~\cite{ashik2022recognizing} curated BTSD, whose images span occlusion, blur, varying distance, and illumination, and benchmarked several state-of-the-art detectors on it. On the same dataset, Rahman \textit{et al.}~\cite{rahman2026fused} addressed the scarcity of annotated data with FUSED-Net, a few-shot detector built on the two-stage Faster R-CNN. Accurately detecting small, occluded, and visually degraded signs under diverse conditions therefore remains open, motivating lightweight architectures that exploit multi-scale representations while maintaining real-time inference.

\section{Methodology}
\begin{figure*}[!t]
\centering
\includegraphics[width=\textwidth]{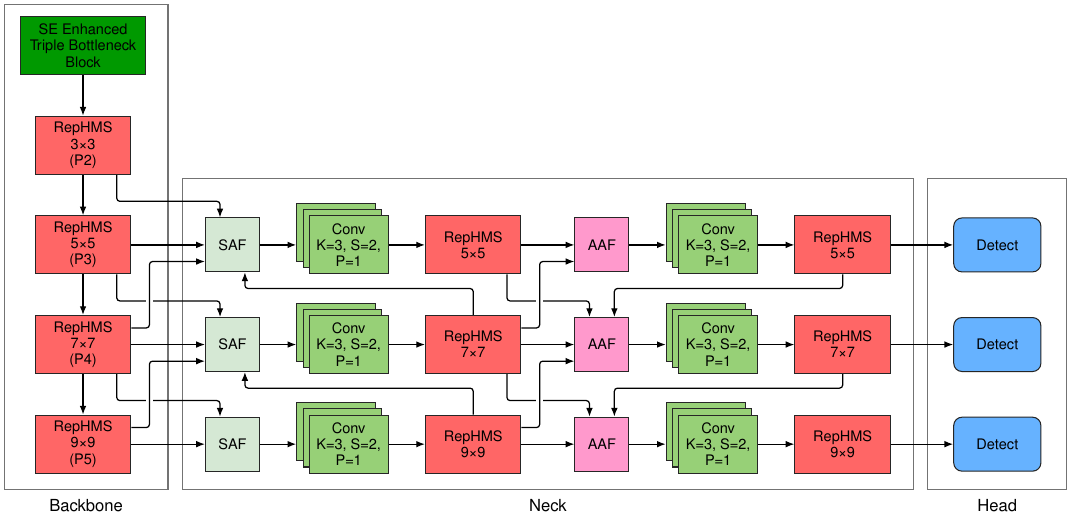}
\caption{Overview of the proposed Squeeze-Aware-YOLO architecture. The model consists of an SE-enhanced input stage with a custom bottleneck block, a hierarchical backbone built using RepHMS modules with multi-scale receptive fields, and a multi-branch neck incorporating Superficial Assisted Fusion (SAF) and Advanced Assisted Fusion (AAF) for effective feature aggregation.}
\label{fig1}
\end{figure*}
Squeeze-Aware-YOLO builds on YOLOv10, which eliminates non-maximum suppression through a dual-assignment training strategy while using large-kernel convolutions and partial self-attention to balance accuracy and efficiency~\cite{wang2024yolov10}.
YOLOv10 nonetheless performs limited early-stage feature refinement: informative channel responses and fine-grained spatial cues are not explicitly emphasized before backbone processing, so weak features from small or visually degraded signs can be suppressed during extraction; in other domains, strengthening YOLOv10's multi-scale feature extraction has reduced missed detections of small targets on cluttered backgrounds~\cite{haoyan2025ead}.

For multi-scale representation we adopt the Re-parameterized Heterogeneous Multi-Scale Module (RepHMS) and multi-branch fusion strategy of MHAF-YOLO~\cite{yang2025mhaf}, whose Multi-Branch Auxiliary Feature Pyramid Network (MAFPN)~\cite{yang2024maf} improves interaction between shallow spatial and deep semantic features but focuses on intermediate and deep interactions without refining low-level features at the input.
We therefore introduce an SE-enhanced triple bottleneck block at the input stage, combining channel-wise attention with heterogeneous multi-scale convolutions to refine features before backbone processing, while the integrated RepHMS modules enrich multi-scale fusion.

\subsection{Architecture Overview}\label{subsec:architecture}
As illustrated in Fig.~\ref{fig1}, the proposed network extends the baseline YOLOv10 architecture through three tightly integrated components that collectively improve hierarchical feature extraction, multi-scale information aggregation, and scale-aware feature fusion.
\begin{enumerate}
\item \textbf{SE-Enhanced Input Stage:} Input images first pass through the SE-enhanced triple bottleneck block, which performs adaptive channel recalibration and multi-scale extraction so that later backbone layers focus on task-relevant features.
\item \textbf{Hierarchical Backbone:} The backbone then extracts hierarchical features through four stages (P2--P5), with RepHMS modules of heterogeneous kernel sizes ($3\times3$, $5\times5$, $7\times7$, and $9\times9$) capturing both local and global context.
\item \textbf{Multi-Branch Feature Pyramid:} The neck combines Superficial Assisted Fusion (SAF), Advanced Assisted Fusion (AAF), and RepHMS modules, integrating high-level semantics with low-level spatial detail for scale-aware aggregation across detection layers.
\end{enumerate}

\subsection{Squeeze-and-Excitation Mechanism}\label{subsec:se}
The SE block~\cite{hu2018squeeze} explicitly models channel inter-dependencies through adaptive recalibration.
For feature tensor $X \in \mathbb{R}^{B \times C \times H \times W}$, where $B$ is batch size, $C$ is channels, and $H \times W$ is spatial resolution, the SE operation consists of the following steps.

\noindent \textbf{Squeeze:} Global average pooling aggregates spatial information per channel:
\begin{equation}
z_c = \frac{1}{HW}\sum_{i=1}^{H} \sum_{j=1}^{W} X_c(i,j), \quad z \in \mathbb{R}^{C}
\label{eq:squeeze}
\end{equation}

\noindent \textbf{Excitation:} Nonlinear channel interactions through bottleneck:
\begin{equation}
s = \sigma(W_{2}\, \delta(W_{1} z))
\label{eq:excitation}
\end{equation}
where $W_1 \in \mathbb{R}^{\frac{C}{r} \times C}$ reduces dimensionality by ratio $r$, $W_2 \in \mathbb{R}^{C \times \frac{C}{r}}$ restores it, $\delta(\cdot)$ is the SiLU activation, and $\sigma(\cdot)$ is sigmoid ensuring $s \in (0,1)^C$.
The resulting $s \in \mathbb{R}^{B \times C \times 1 \times 1}$ is broadcast across the $H \times W$ spatial grid in~\eqref{eq:recalibration}.

\noindent \textbf{Recalibration:} Channel-wise multiplication applies learned weights:
\begin{equation}
\tilde{X}_{c} = s_{c} \cdot X_{c}, \quad c = 1,\ldots,C
\label{eq:recalibration}
\end{equation}

Computational complexity is $\mathcal{O}(\frac{2C^2}{r})$, negligible for typical reduction ratios ($r = 8$ or $16$) compared to convolutional costs.

\subsection{SE-Enhanced Triple Bottleneck Block}\label{subsec:se_bottleneck}
\begin{figure}[!ht]
    \centering
    \includegraphics[width=0.5\linewidth]{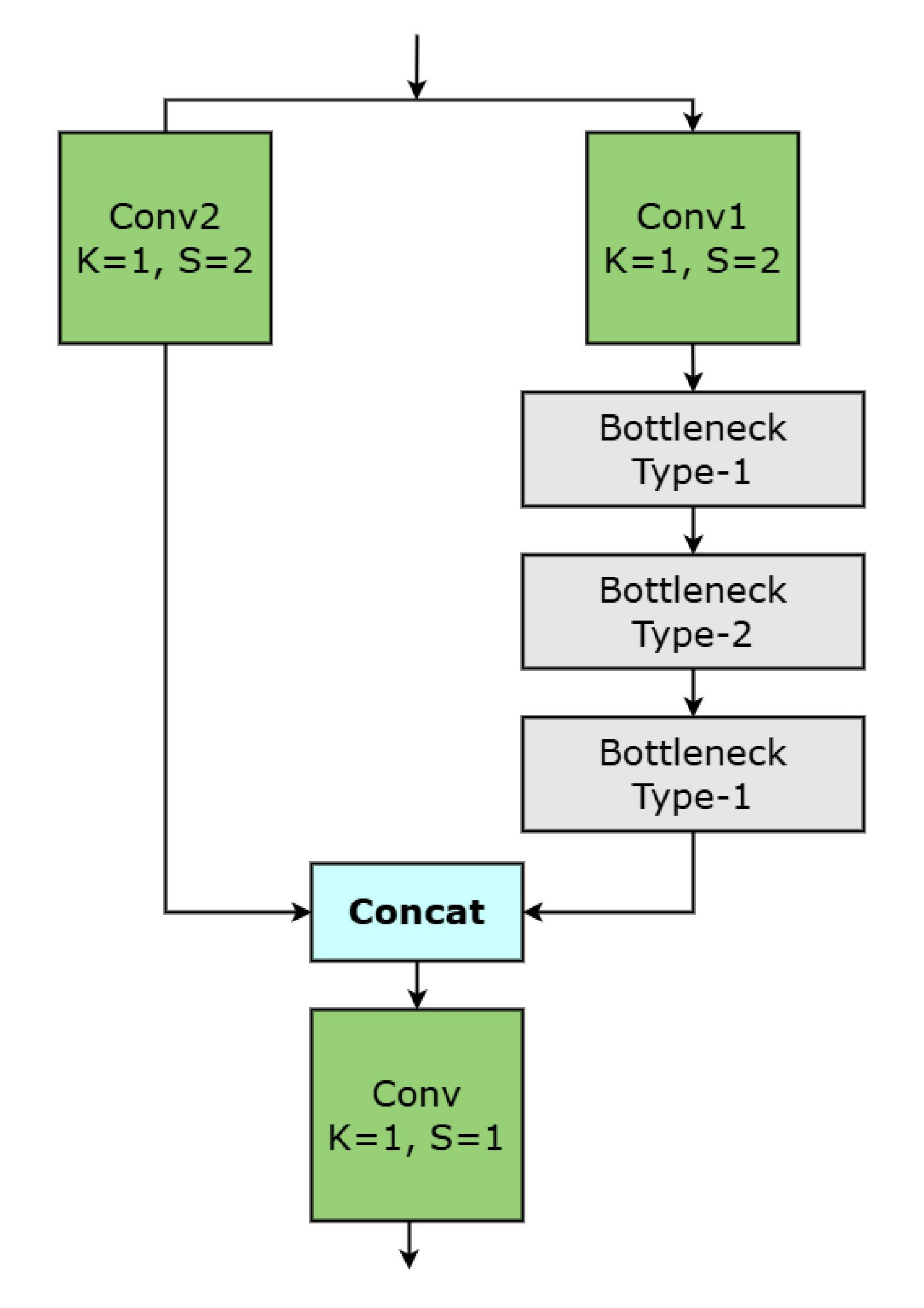}
    \caption{Simplified view of the SE-enhanced bottleneck block, showing the dual-path structure with sequential bottleneck processing and a compressed bypass path.}
    \label{fig2}
\end{figure}

\begin{figure}[!ht]
\centering
\subfloat[Bottleneck Type-2 with $5 \times 5$ convolution]{\includegraphics[width=0.48\linewidth,page=1]{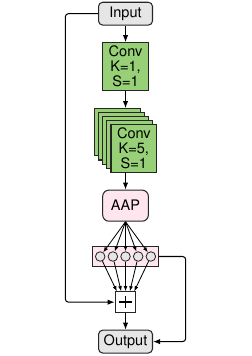}\label{fig3A}}
\hfil
\subfloat[Bottleneck Type-1 with $3 \times 3$ convolution]{\includegraphics[width=0.48\linewidth,page=2]{figures/fig3_bottlenecks.pdf}\label{fig3B}}
\caption{Two bottleneck variants used in the proposed architecture. (a) Bottleneck Type-2 employs a $5 \times 5$ convolution, whereas (b) Bottleneck Type-1 uses a $3 \times 3$ convolution.}
\label{fig3}
\end{figure}

\begin{figure}[!ht]
\centering
\includegraphics[width=\linewidth]{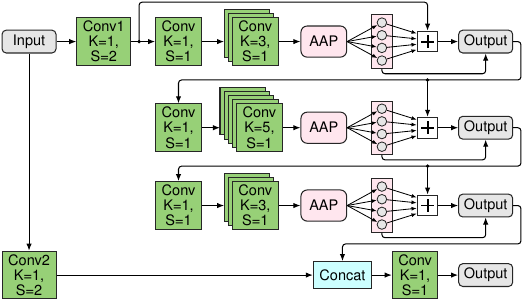}
\caption{Detailed SE-enhanced bottleneck block, showing the dual-path structure: the main path processes through three sequential bottlenecks (Type-1, Type-2, Type-1) with varying kernels, while the bypass path provides direct gradient flow.}
\label{fig4}
\end{figure}

We introduce an SE-enhanced bottleneck incorporating multi-scale feature extraction and adaptive channel recalibration at the network entry.
Fig.~\ref{fig2} presents a simplified overview, while Fig.~\ref{fig4} shows complete architectural details.
As illustrated in Fig.~\ref{fig2}, the input feature map is divided into two parallel branches.
The first branch performs downsampling using a convolutional layer with a $1 \times 1$ kernel and stride of 2, preserving the shortcut information.
The second branch applies the same convolutional operation, followed by the proposed triple bottleneck module, where a Type-2 bottleneck is sandwiched between two Type-1 bottlenecks.
This dual-branch design enables simultaneous preservation of coarse features and progressive refinement of discriminative representations before feature aggregation.
The architecture employs three sequential bottlenecks of two types (Fig.~\ref{fig3}):

\noindent \textbf{Bottleneck Type-1:} Uses $1\times1$ convolution for channel projection, followed by $3\times3$ and depthwise convolutions for local features. Adaptive Average Pooling (AAP) ensures scale invariance, a fully connected layer generates compact descriptors, and skip connections aid gradient flow.

\noindent \textbf{Bottleneck Type-2:} Employs $5 \times 5$ convolution for larger patterns for medium-to-large features. The flow mirrors Type-1 but with an expanded receptive field.

Fig.~\ref{fig4} illustrates the complete block by integrating all previously described components into a unified architecture.
The primary path processes input sequentially: (1) Bottleneck Type-1 for fine-grained local details, (2) Bottleneck Type-2 expanding the receptive field for broader context, and (3) another Bottleneck Type-1 refining features locally.
Each bottleneck incorporates Adaptive Average Pooling (AAP) and Fully Connected (FC) layers with skip connections.
The second path serves as a compressed bypass, providing gradient stability and direct information flow.
Both paths are concatenated channel-wise and projected through a final $1 \times 1$ convolution.
Concretely, each branch maps $X \in \mathbb{R}^{B \times C_{\mathrm{in}} \times H \times W}$ to $\mathbb{R}^{B \times C_{\mathrm{out}}/2 \times H/2 \times W/2}$ via a stride-2 $1 \times 1$ convolution; the bottlenecks preserve this shape, channel-axis concatenation gives $\mathbb{R}^{B \times C_{\mathrm{out}} \times H/2 \times W/2}$, and the final $1 \times 1$ convolution projects back to $C_{\mathrm{out}}$.
This early placement enables channel recalibration and multi-scale extraction before hierarchical abstraction in deeper layers.

\subsection{Multi-Scale Feature Fusion}
The Multi-Branch Auxiliary FPN (MAFPN)~\cite{yang2024maf,yang2025mhaf} addresses standard FPN limitations through bidirectional and multi-path connections preserving fine-grained spatial cues and high-level semantic abstractions.

\noindent \textbf{Superficial Assisted Fusion (SAF):} Integrates deep information with shallow spatial features for small object detection. Indexing pyramid levels so that $P_{n} \in \mathbb{R}^{H_{n} \times W_{n} \times C_{n}}$ with $H_{n+1} = H_{n}/2$ and $W_{n+1} = W_{n}/2$, SAF aligns its inputs to the resolution of $P_{n}$ before fusion:
\begin{equation}
\begin{aligned}
P'_{n} = \operatorname{concat}\Big(
    &\delta(\operatorname{GAP}(P_{n-1})),\;
    P_{n}, \\
    &\delta(\operatorname{Conv}(U(P_{n+1}))),\;
    U(P'_{n+1})
\Big)
\end{aligned}
\label{eq:saf}
\end{equation}

\noindent \textbf{Advanced Assisted Fusion (AAF):} Aggregates information from shallow high-resolution, shallow low-resolution, sibling, and previous layers:
\begin{equation}
\begin{aligned}
P''_{n} = \operatorname{concat}\Big(
    &\delta(\operatorname{Down}(P'_{n-1})),
    \delta(\operatorname{GAP}(P''_{n-1})), \\
    &P'_{n},
    \delta(\operatorname{Conv}(U(P'_{n+1})))
\Big)
\end{aligned}
\label{eq:aaf}
\end{equation}
In both~\eqref{eq:saf} and~\eqref{eq:aaf}, $U(\cdot)$ denotes nearest-neighbour upsampling and $\operatorname{Down}(\cdot)$ strided convolutional downsampling, each resampling its input to the target resolution $H_{n} \times W_{n}$; $\operatorname{GAP}$ denotes global average pooling, whose $1 \times 1$ output is broadcast across the $H_{n} \times W_{n}$ grid; $\operatorname{Conv}$ denotes a $1 \times 1$ convolution projecting its branch to a common channel width; and $\delta$ is the SiLU activation.
Every argument of $\operatorname{concat}$ is therefore spatially aligned to $H_{n} \times W_{n}$ beforehand, concatenation is performed along the channel axis, and a subsequent $1 \times 1$ convolution projects the concatenated tensor back to $C_{n}$ channels, enabling multi-layer feature integration for scale-aware detection.

\noindent \textbf{Re-parameterized Heterogeneous Multi-Scale Module (RepHMS):}
RepHMS~\cite{yang2025mhaf} enhances multi-scale feature representation at low computational cost.
The input is projected through a $1 \times 1$ convolution and split into $N$ parallel branches: the first preserves the original features, while the remaining branches progressively refine them using $M$ cascaded \textit{Re-parameterized Heterogeneous Depthwise Convolution (RepHDWConv)} blocks, each branch feeding the next to promote feature reuse and gradient flow.
All branch outputs are concatenated along the channel axis and fused by a $1 \times 1$ convolution, with $M$ and $N$ controlling module depth and width.
RepHDWConv applies depthwise convolutions of several kernel sizes during training to enlarge the effective receptive field, and re-parameterizes them into a single depthwise convolution at inference, retaining the multi-scale benefit at no additional inference cost.
This strengthens the interaction between low-level spatial detail and high-level semantics throughout the backbone and neck.

\section{Experiments}\label{sec:experiments}
\subsection{Datasets}\label{sec:datasets}
We utilized three publicly available traffic sign detection datasets: the CSUST Chinese Traffic Sign Detection Benchmark 2021~\cite{zhang2022cctsdb}, the Bangladeshi Traffic Sign Dataset~\cite{ashik2022recognizing}, and the German Traffic Sign Detection Benchmark~\cite{houben2013detection}.

\noindent \textbf{CSUST Chinese Traffic Sign Detection Benchmark 2021 (CCTSDB2021):}
Contains 18,356 traffic scene images covering mandatory, prohibitory, and warning signs, including 5,260 additional real-world images to increase scene diversity. The dataset is balanced across five sign-size levels and six weather conditions, with a dedicated 2,000-image test set containing negative samples for robust evaluation. The training subset consists of 16,356 images~\cite{zhang2022cctsdb}.

\noindent \textbf{Bangladeshi Traffic Sign Dataset (BTSD):}
Comprises 2,986 images containing 3,063 annotated traffic signs across 15 classes. The dataset captures challenging real-world Bangladeshi road conditions, including cluttered traffic scenes, occlusions, and motion blur. The training set contains 2,324 images, while the test set contains 662 images~\cite{ashik2022recognizing}.

\noindent \textbf{German Traffic Sign Detection Benchmark (GTSDB):}
Comprises 900 road images with approximately 1,200 annotated traffic signs from Prohibitory, Danger, and Mandatory categories~\cite{houben2013detection}. Compared with the other datasets, GTSDB contains relatively less complex traffic scenes. We use an extended version with 296 additional images, consisting of 788 training images, 218 validation images, and 190 test instances.

\noindent \textbf{Evaluation scope:}
Each dataset is trained and evaluated independently, so our results establish multi-benchmark consistency rather than cross-domain transfer.
The three benchmarks use incompatible label spaces---BTSD annotates 15 fine-grained sign classes, whereas CCTSDB2021 and GTSDB annotate three and four super-classes respectively---so a single model cannot be evaluated across them without a label-space unification that alters the detection task itself.
We therefore treat cross-dataset transfer as a separate study and report it as a limitation of this work.

\subsection{Experimental Setup}\label{sec:experimental_setup}
All experiments used an NVIDIA GeForce RTX 3090 GPU with Python 3.12.12, Ultralytics 8.1.34, PyTorch 2.2.0 (CUDA 12.1), and NumPy 1.26.4.
All models were trained for up to 700 epochs at $640 \times 640$ input resolution with batch size 16 and early stopping after 70 epochs without improvement.
The optimizer was selected automatically by Ultralytics (\texttt{optimizer=auto}), which resolves to SGD at the training lengths used here, with momentum $0.937$ and weight decay $5\times10^{-4}$.
The learning rate started at $0.01$ and decayed to $1\%$ of that value, preceded by a three-epoch warmup (warmup momentum $0.8$, bias learning rate $0.1$).
Augmentation comprised mosaic (probability $1.0$, disabled for the final ten epochs), random scaling ($0.9$), translation ($0.1$), horizontal flipping ($0.5$), random erasing ($0.4$), RandAugment, and HSV jitter (hue $0.015$, saturation $0.7$, value $0.4$); mixup, vertical flipping, rotation, shear, and perspective were disabled.
Weights were initialized from pretrained MAF-YOLOv2-N parameters~\cite{yang2025mhaf}, of which $1{,}399$ of $1{,}411$ parameter groups transferred, with the remaining groups randomly initialized.
Preprocessing followed the Ultralytics pipeline: images letterboxed to $640 \times 640$ with aspect ratio preserved, pixel values scaled to $[0,1]$, and annotations in normalized YOLO format, with no class re-mapping, re-balancing, or image filtering.
All three datasets were trained with this identical configuration and no per-dataset hyperparameter tuning.

\noindent \textbf{Evaluation Metrics:}
We report Precision, Recall, F1-score, per-class Average Precision (AP, the area under the precision--recall curve), and mean Average Precision (mAP) averaged over classes at IoU $0.5$.

\section{Results and Discussion}\label{sec:results_discussion}
\subsection{Overall Performance}
We compared our approach against three baseline YOLO models (YOLOv8~\cite{jocher2023yolov8}, YOLOv10~\cite{wang2024yolov10}, YOLOv11~\cite{jocher2024yolo11}) and a recent multi-branch architecture, MHAF-YOLO~\cite{yang2025mhaf}.
Despite incorporating triple bottlenecks with SE modules, our Squeeze-Aware-YOLO maintains a comparable parameter count (2.2M).
Table~\ref{tab:model_params_overall_map} summarizes mAP@IoU=0.5 across datasets.
\begin{table}[htbp]
\caption{mAP@0.5 Comparison Across Datasets}
\begin{center}
\begin{tabular}{|l|c|c|c|}
\hline
\textbf{Model} & \textbf{\textit{BTSD}} & \textbf{\textit{CCTSDB2021}} & \textbf{\textit{GTSDB}} \\
\hline
YOLOv8 & 0.663 & 0.794 & 0.809 \\
\hline
YOLOv10 & 0.679 & 0.792 & 0.826 \\
\hline
YOLOv11 & 0.668 & \textbf{0.804} & 0.836 \\
\hline
MHAF-YOLO & 0.685 & 0.769 & 0.816 \\
\hline
\textbf{Squeeze-Aware-YOLO} & \textbf{0.747} & 0.794 & \textbf{0.840} \\
\hline
\end{tabular}
\label{tab:model_params_overall_map}
\end{center}
\end{table}

\subsection{Dataset-Specific Analysis}
\subsubsection{Bangladeshi Traffic Sign Dataset (BTSD)}
Our model achieved 74.7\% mAP@0.5, a 6.2 percentage point increase over the next best model, demonstrating the effectiveness of the SE-enhanced bottleneck in emphasizing informative channels and suppressing background noise.
Table~\ref{tab:class_wise_model_map_bd} reports class-wise performance: the proposed model attains the best score on five of the 15 classes and the highest overall mAP, with gains concentrated on classes the baselines handle poorly and regressions on classes they already solve, as analyzed in Section~\ref{sec:results_discussion}.
\begin{table}[!t]
\caption{Class-Wise mAP@0.5 on the Bangladeshi Traffic Sign Detection Dataset}
\begin{center}
\footnotesize
\setlength{\tabcolsep}{2.5pt}
\begin{tabular}{|>{\raggedright\arraybackslash}p{1.75cm}|c|c|c|c|c|}
\hline
\textbf{Class} &
\textbf{YOLOv8} &
\textbf{YOLOv10} &
\textbf{YOLOv11} &
\textbf{MHAF} &
\textbf{Ours} \\
\hline
Ped. Crossing & \textbf{0.760} & 0.735 & 0.425 & 0.675 & 0.730 \\
\hline
SL40          & 0.845 & 0.775 & 0.722 & 0.759 & \textbf{0.850} \\
\hline
SL80          & 0.563 & \textbf{0.729} & 0.725 & 0.709 & 0.705 \\
\hline
SL60          & 0.622 & 0.671 & 0.737 & 0.894 & \textbf{0.950} \\
\hline
No O/T        & \textbf{0.964} & 0.915 & 0.947 & 0.906 & 0.958 \\
\hline
Bend (R)      & 0.794 & 0.821 & 0.793 & 0.687 & \textbf{0.894} \\
\hline
Bend (L)      & 0.865 & \textbf{0.915} & 0.858 & 0.185 & 0.320 \\
\hline
Side Rd. (R)  & 0.180 & 0.302 & 0.379 & \textbf{0.736} & 0.627 \\
\hline
Narrow Bridge & 0.568 & \textbf{0.740} & 0.544 & 0.558 & 0.638 \\
\hline
No Horn       & 0.331 & 0.536 & 0.428 & 0.840 & \textbf{0.952} \\
\hline
Road Hump     & 0.887 & 0.644 & \textbf{0.902} & 0.513 & 0.687 \\
\hline
School        & 0.432 & 0.385 & 0.404 & \textbf{0.911} & 0.883 \\
\hline
Side Rd. (L)  & 0.421 & 0.399 & 0.509 & 0.758 & \textbf{0.831} \\
\hline
U-Turn        & \textbf{0.938} & 0.933 & 0.916 & 0.529 & 0.514 \\
\hline
Crossroads    & \textbf{0.770} & 0.645 & 0.728 & 0.622 & 0.660 \\
\hline
\textbf{Overall} & 0.663 & 0.679 & 0.668 & 0.685 & \textbf{0.747} \\
\hline
\multicolumn{6}{p{8.1cm}}{\footnotesize $^{\mathrm{a}}$Abbreviations: Ped. = Pedestrian, SL = Speed Limit, O/T = Overtaking, Rd. = Road, Bend (R) = Sharp Bend Right, Bend (L) = Sharp Bend Left, MHAF = MHAF-YOLO. Results are reported on the BTSD test set.}
\end{tabular}
\label{tab:class_wise_model_map_bd}
\end{center}
\end{table}

\subsubsection{CCTSDB2021 Dataset}
CCTSDB2021 poses an easier detection problem than BTSD: signs are larger and less occluded, its three sign types are visually distinctive, and the class distribution is more balanced.
Here YOLOv11 achieves the highest overall mAP (0.804) and leads every category, while our model places second alongside YOLOv8 and matches YOLOv11 on the Prohibitory class (Table~\ref{tab:class_map_val_cctsdb}).
Early feature refinement nonetheless helps, as our architecture consistently beats MHAF-YOLO across categories.
\begin{table}[!ht]
\caption{Class-Wise mAP@0.5 on CCTSDB2021}
\begin{center}
\renewcommand{\arraystretch}{1.15}
\begin{tabular}{|l|c|c|c|c|}
\hline
\textbf{Model} & \textbf{Prohibitory} & \textbf{Mandatory} & \textbf{Warning} & \textbf{Overall} \\
\hline
YOLOv8      & 0.798 & 0.737 & 0.847 & 0.794 \\
\hline
YOLOv10     & 0.794 & 0.728 & 0.855 & 0.792 \\
\hline
YOLOv11     & \textbf{0.810} & \textbf{0.739} & \textbf{0.864} & \textbf{0.804} \\
\hline
MHAF-YOLO   & 0.782 & 0.687 & 0.838 & 0.769 \\
\hline
\textbf{Ours} & \textbf{0.810} & 0.712 & 0.858 & 0.794 \\
\hline
\end{tabular}
\label{tab:class_map_val_cctsdb}
\end{center}
\end{table}

\subsubsection{German Traffic Sign Detection Benchmark (GTSDB)}
On GTSDB---which contains diverse signs but lacks the extreme clutter and occlusion of BTSD---our model achieves the highest mAP of 0.840.
Table~\ref{tab:class_map_val_gtsdb} shows it leads the Mandatory category (0.760) and ties YOLOv11 on Others (0.825), while trailing narrowly behind YOLOv8 on Danger and YOLOv10 on Prohibitory.
\begin{table}[!ht]
\caption{Class-Wise mAP@0.5 on GTSDB}
\begin{center}
\footnotesize
\setlength{\tabcolsep}{4pt}
\begin{tabular}{|l|c|c|c|c|c|}
\hline
\textbf{Model} & \textbf{Prohibitory} & \textbf{Mandatory} & \textbf{Danger} & \textbf{Others} & \textbf{Overall} \\
\hline
YOLOv8      & 0.799 & 0.711 & \textbf{0.949} & 0.636 & 0.809 \\
\hline
YOLOv10     & \textbf{0.919} & 0.726 & 0.906 & 0.755 & 0.826 \\
\hline
YOLOv11     & 0.907 & 0.733 & 0.878 & \textbf{0.825} & 0.836 \\
\hline
MHAF-YOLO   & 0.900 & 0.730 & 0.866 & 0.767 & 0.816 \\
\hline
\textbf{Ours} & 0.910 & \textbf{0.760} & 0.865 & \textbf{0.825} & \textbf{0.840} \\
\hline
\end{tabular}
\label{tab:class_map_val_gtsdb}
\end{center}
\end{table}

\subsection{Qualitative and Failure Analysis}
\begin{figure*}[!t]
\centering
\includegraphics[width=\textwidth]{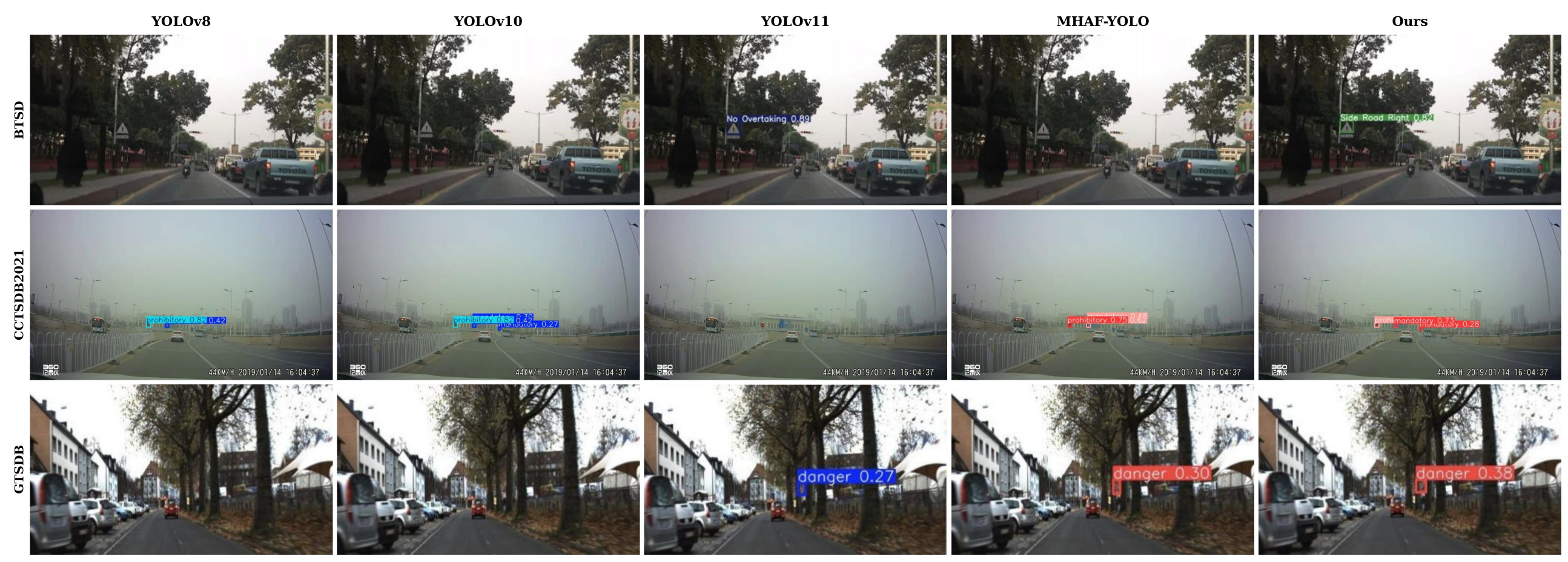}
\caption{Qualitative detection comparison across three traffic-sign datasets. Rows correspond to the BTSD, CCTSDB2021, and GTSDB datasets, and columns show results from YOLOv8, YOLOv10, YOLOv11, MHAF-YOLO, and our method.}
\label{fig5}
\end{figure*}
Fig.~\ref{fig5} compares YOLOv8, YOLOv10, YOLOv11, MHAF-YOLO, and the proposed Squeeze-Aware-YOLO on BTSD, CCTSDB2021, and GTSDB scenes covering background clutter, small objects, and partial occlusion.
In the first row, a BTSD sign heavily occluded by vegetation is missed by YOLOv8, YOLOv10, and MHAF-YOLO and misclassified by YOLOv11, whereas the proposed model localizes and classifies it correctly.
The second row shows a cluttered CCTSDB2021 scene with multiple small, partially occluded signs: YOLOv11 detects none, YOLOv8 two, and YOLOv10 three with low confidence, while MHAF-YOLO and the proposed model detect all targets, the latter with higher confidence.
The third row contains a GTSDB sign occluded by tree branches, missed by YOLOv8 and YOLOv10, misdetected by YOLOv11, and detected with low confidence by MHAF-YOLO, whereas the proposed model detects it confidently.
These are selected successes rather than a representative sample, so we complement them with a systematic analysis of where the model fails.

\noindent \textbf{Where the gains and losses fall.}
Grouping the 15 BTSD classes of Table~\ref{tab:class_wise_model_map_bd} by the strength of the best YOLO baseline reveals a systematic trade-off (Table~\ref{tab:failure_bands}).
On the four classes the baselines handle worst the proposed model gains $35.9$ mAP points on average and improves on all four; on the seven mid-difficulty classes it is essentially unchanged; and on the four the baselines already solve well it loses $31.0$ points and regresses on all four.
The method therefore redistributes capacity toward rare and visually degraded signs at a measurable cost on easy, well-separated ones; overall mAP still improves because the weak classes carry most of the available headroom.
\begin{table}[!t]
\caption{Per-Class mAP@0.5 Change ($\Delta$) on BTSD Relative to the Best of YOLOv8, YOLOv10, and YOLOv11, Grouped by That Baseline's Difficulty}
\begin{center}
\footnotesize
\setlength{\tabcolsep}{3pt}
\begin{tabular}{|l|c|c|c|}
\hline
\textbf{Baseline band} & \textbf{Classes} & \textbf{Mean $\Delta$mAP} & \textbf{Improved} \\
\hline
Hard ($<0.60$) & 4 & $+0.359$ & 4/4 \\
\hline
Moderate ($\geq 0.60$, $<0.90$) & 7 & $+0.004$ & 3/7 \\
\hline
Easy ($\geq 0.90$) & 4 & $-0.310$ & 0/4 \\
\hline
\end{tabular}
\label{tab:failure_bands}
\end{center}
\end{table}

\noindent \textbf{The two largest regressions.}
Sharp Bend (Left) falls to $0.320$ while its near-mirror-image counterpart Sharp Bend (Right) reaches $0.894$, and U-Turn falls to $0.514$.
Both failures are inherited rather than introduced: MHAF-YOLO, the backbone and neck we build on, scores $0.185$ and $0.529$ on these two classes, and the proposed modules recover part of the gap, improving on $11$ of $15$ classes over MHAF-YOLO.
A likely contributing factor for the mirrored pair is that squeeze-and-excitation recalibrates channels and is orientation-agnostic, so it cannot by itself disambiguate mirror-symmetric shapes, while the horizontal-flip augmentation used during training maps a left-bend sign onto a right-bend appearance without changing its label.
Orientation-sensitive augmentation policies and spatially selective attention are therefore a necessary direction for mirror-symmetric sign classes, and we regard the present model as unsuitable for deployment on such classes without that correction.

\subsection{Ablation Study}
To evaluate the contributions of each proposed component, we conducted an ablation study on the three benchmark datasets.
Starting from the baseline MHAF-YOLO model, we incrementally introduce architectural components to analyze their individual and combined effects.
All experiments use identical training conditions for fair comparison.
The Triple Layer Bottleneck (TLB) is first integrated with MHAF, followed by joint incorporation of both TLB and Squeeze-and-Excitation modules to form the complete Squeeze-Aware-YOLO.
Table~\ref{table:ablation_study} reports results for all configurations.
\begin{table}[!t]
\caption{Ablation Study Showing mAP@0.5 for Different Module Configurations on the BTSD, CCTSDB2021, and GTSDB Datasets}
\begin{center}
\setlength{\tabcolsep}{5pt}
\begin{tabular}{|c|c|c|c|c|c|}
\hline
\multicolumn{3}{|c|}{\textbf{Components}} &
\multicolumn{3}{c|}{\textbf{mAP@0.5}} \\
\hline
\textbf{MHAF Baseline} & \textbf{TLB} & \textbf{SE} &
\textbf{BTSD} & \textbf{CCTSDB2021} & \textbf{GTSDB} \\
\hline
\cmark & \xmark & \xmark & 0.685 & 0.769 & 0.816 \\
\hline
\cmark & \cmark & \xmark & 0.722 & 0.783 & 0.832 \\
\hline
\cmark & \cmark & \cmark & \textbf{0.747} & \textbf{0.794} & \textbf{0.840} \\
\hline
\multicolumn{6}{p{8.1cm}}{\footnotesize $^{\mathrm{a}}$TLB denotes the Triple Layer Bottleneck module, and SE denotes the Squeeze-and-Excitation module. Values in the final row correspond to the complete proposed configuration (MHAF + TLB + SE).}
\end{tabular}
\label{table:ablation_study}
\end{center}
\end{table}
Compared to the baseline MHAF-YOLO, adding TLB yields mAP gains of 3.7 percentage points on BTSD, 1.4 on CCTSDB2021, and 1.6 on GTSDB, indicating that TLB improves feature extraction through deeper and more efficient transformations.
Integrating both SE and TLB achieves the highest performance on all datasets, with gains of 6.2, 2.5, and 2.4 percentage points over the baseline respectively.
Additional gains over the TLB-only configuration demonstrate the effectiveness of the SE module in recalibrating channel-wise feature responses.
Notably, both components improve mAP in the same direction on all three datasets under an identical training configuration with no per-dataset tuning, indicating that the architectural benefit is consistent across domains of differing difficulty, even though cross-domain transfer of the learned weights remains untested.
\begin{figure}[!ht]
\centering
\includegraphics[width=\linewidth]{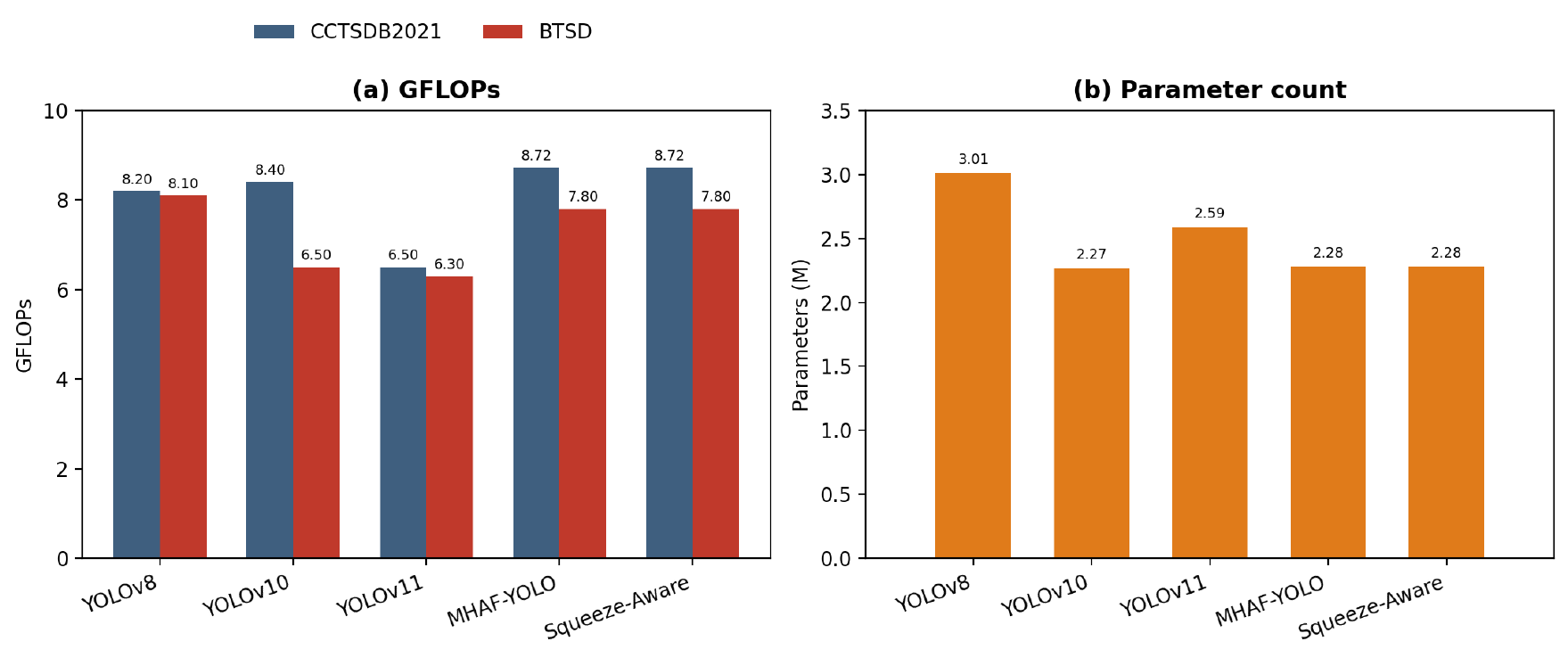}
\caption{Computational complexity of the evaluated models on CCTSDB2021 and BTSD: (a) GFLOPs for model and (b) parameter count (millions).}
\label{fig6}
\end{figure}

\begin{figure}[!ht]
\centering
\includegraphics[width=\linewidth]{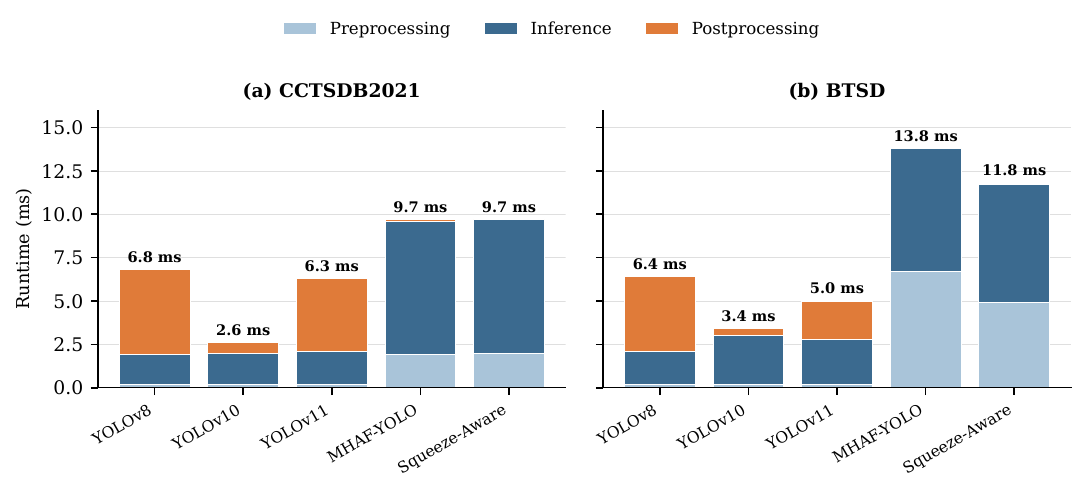}
\caption{Per-stage runtime decomposition (preprocessing, inference, postprocessing) on (a) CCTSDB2021 and (b) BTSD. Labels above each bar report the total end-to-end latency.}
\label{fig7}
\end{figure}

Figs.~\ref{fig6} and~\ref{fig7} compare computational complexity and runtime on the CCTSDB2021 and Bangladeshi (BTSD) datasets.
All models are of similar complexity, with GFLOPs between 6.3 and 8.7 and parameter counts between 2.27M and 3.01M, so runtime is governed more by pipeline efficiency than by model size; despite the additional squeeze-aware module, Squeeze-Aware-YOLO remains lightweight.
YOLOv10 attains the lowest latency on both datasets owing to its minimal postprocessing overhead, while YOLOv8 and YOLOv11 offer competitive inference speed but higher postprocessing cost.
MHAF-YOLO and Squeeze-Aware-YOLO share identical computational complexity, yet MHAF-YOLO incurs higher preprocessing latency, resulting in longer execution times.
Squeeze-Aware-YOLO achieves a balanced runtime profile, requiring 9.7 ms on CCTSDB2021 and 11.8 ms on BTSD.
Although slightly slower than the standard YOLO variants, this modest overhead yields more effective feature extraction, improving detection accuracy and robustness in challenging traffic scenes.

\section{Conclusion}\label{sec:conclusion}
We introduced Squeeze-Aware-YOLO, an architecture for traffic sign detection that addresses varying lighting conditions, viewing distances, and cluttered backgrounds. Its SE-enhanced triple bottleneck block performs early-stage feature refinement through adaptive channel recalibration and multi-scale feature extraction, yielding more discriminative representations.
Across three benchmarks the model achieves 74.7\% mAP@0.5 on BTSD, 79.4\% on CCTSDB2021, and 84.0\% on GTSDB with only 2.2M parameters, showing that early-stage channel recalibration combined with heterogeneous kernel selection balances detection accuracy against computational cost. The architecture performs consistently across benchmarks with large variations in traffic sign scale, image resolution, and scene complexity, and its lightweight design suits real-time inference on embedded and resource-constrained platforms.
Our protocol trains and evaluates each dataset independently, so it establishes multi-benchmark consistency rather than cross-domain transfer, and the per-class analysis in Section~\ref{sec:results_discussion} shows that the gains concentrate on classes the baselines handle poorly. In future work we plan to investigate asymmetric convolutional designs and lightweight attention mechanisms within the detection head, and to quantify cross-domain generalization directly by unifying the benchmarks under a shared super-class label space (prohibitory, mandatory, warning) and a class-agnostic protocol, enabling train-on-one, test-on-another evaluation across all dataset pairs.

\bibliographystyle{IEEEtran}
\bibliography{IEEEexample}

\end{document}